\documentclass[pdflatex,sn-mathphys-num]{sn-jnl}

\usepackage[utf8]{inputenc}
\usepackage{graphicx}%
\usepackage{multirow}%
\usepackage{amsmath,amssymb,amsfonts}%
\usepackage{amsthm}%
\usepackage{mathrsfs}%
\usepackage[title]{appendix}%
\usepackage{xcolor}%
\usepackage{textcomp}%
\usepackage{manyfoot}%
\usepackage{booktabs}%
\usepackage{physics}
\usepackage{braket}
\usepackage{newunicodechar}
\newunicodechar{č}{\v{c}}

\begin{document}

\title[A hybrid quantum-classical neural network for learning to route]{A hybrid quantum-classical neural network for learning to route}

\author*[1,3]{\fnm{Marcus Rolf Peter} \sur{Ritt}}\email{marcus.ritt.BE@eldorado.org.br}

\author[1]{\fnm{Alexsandro Santos} \sur{da Rosa J\'unior}}\email{alexsandro.junior@eldorado.org.br}

\author[1]{\fnm{Marcos Vinicius} \sur{Reballo}}\email{marcos.reballo@eldorado.org.br}

\author[1,2]{\fnm{Cesar Augusto} \sur{do Amaral}}\email{cesar.amaral.BE@eldorado.org.br}

\author[1]{\fnm{Fernando Augusto Caletti} \sur{de Barros}}\email{fernando.barros@eldorado.org.br}

\affil*[1]{\orgname{Instituto de Pesquisas Eldorado}, \orgaddress{\city{Porto Alegre}, \state{RS}, \country{Brazil}}}

\affil[2]{\orgdiv{Departamento de F\'isica}, \orgname{Universidade Federal de Santa Catarina}, \orgaddress{\city{Florian\'opolis}, \postcode{88040-900}, \state{SC}, \country{Brazil}}}

\affil[3]{\orgdiv{Instituto de Inform\'atica}, \orgname{Universidade Federal do Rio Grande do Sul}, \orgaddress{\city{Porto Alegre}, \country{Brazil}}}

\abstract{
We study hybrid quantum-classical neural networks for learning routing heuristics. Specifically, we ask whether small quantum neural networks can replace parameter-heavy modules inside a well-established attention-based routing model while maintaining solution quality. We consider the capacitated vehicle routing problem (CVRP) and its variant with time windows (CVRPTW), evaluating instances with 10 to 100 customers. We investigate four-qubit quantum neural networks (QNNs) as replacements for encoder feed-forward modules and attention output projections, comparing them against width-matched classical bottlenecks. For these problems, encoder feed-forward replacement emerges as the most promising design, reducing the number of model parameters by 56.6\% relative to the original classical architecture. Across matched comparisons, the QNN bottlenecks retain solution quality close to their classical counterparts, with differences in mean route length of approximately 0.5\% or less under feed-forward replacement. Replacing the attention output projections as well increases parameter reduction to 63.2\%, but leads to larger performance gaps for larger instances. The QNNs do not systematically outperform the classical controls. These results identify encoder feed-forward replacement as a viable hybrid-module compression strategy for neural combinatorial optimization, without demonstrating a quantum-specific advantage.
}

\keywords{Vehicle routing, hybrid quantum-classical neural networks, neural combinatorial optimization, quantum machine learning.}

\maketitle

\section{Introduction}

There is ongoing research on solving combinatorial optimization problems by learning, in particular by training neural networks to solve them from sampled instances instead of designing problem-specific algorithms manually. Current neural architectures for this task, however, have too many parameters to be replaced entirely by quantum neural networks on near-term devices.

This paper studies whether hybrid quantum-classical solutions can achieve good performance with fewer parameters, and thereby point to possible ways of replacing selected components of classical neural networks by quantum versions. This work focuses on the classical capacitated vehicle routing problem (CVRP), a strongly NP-hard routing problem with direct practical relevance~\citep{Dantzig.Ramser/1959}.

Classical approaches to the CVRP range from exact branch-and-cut formulations to a wide variety of constructive heuristics, local-search methods, and metaheuristics~\citep{toth2014vrp,bogyrbayeva2024mlvrp}. In parallel, there is growing interest in quantum optimization methods such as QAOA and VQE~\citep{Farhi.etal/2014a,Peruzzo.etal/2013}, which, when applied to combinatorial optimization, can use QUBO formulations mapped to Ising Hamiltonians 
and solve them as variational hybrid quantum-classical programs. More recently, quantum machine learning (QML) methods have been proposed, in which parametric quantum circuits act as learnable components inside larger models. However, current quantum devices and simulators severely limit circuit width and depth, which makes it difficult to replace entire neural architectures for routing. This work therefore focuses on a narrower question: whether small QNN modules can replace parameter-heavy subcomponents of competitive classical neural models while keeping most of their solution quality.

This paper makes three main contributions. First, it develops hybrid attention models for the CVRP and CVRPTW in which small QNNs replace selected parameter-heavy encoder modules. Second, it compares these QNN modules with parameter-matched classical bottlenecks, separating the effect of quantum processing from that of compression alone. Third, it evaluates the resulting policies alongside established classical routing methods on fixed held-out instances. The single-compression hybrid reduces the model size by more than $56\%$ while retaining solution quality close to its matched classical neural baseline.

The rest of this paper is organized as follows. Section~\ref{sec:vrp} introduces the vehicle routing problem. Section~\ref{sec:related} discusses related work. Section~\ref{sec:hybrid} presents the proposed hybrid quantum-classical neural network. Section~\ref{sec:methodology} describes the experimental methodology. Section~\ref{sec:results} reports the experimental results. Section~\ref{sec:conclusion} concludes the paper.

\section{The Vehicle Routing Problem}\label{sec:vrp}

The CVRP can be defined as follows. Let $V = \{0, 1, \dots, n\}$ be the set of nodes, where $0$ represents the depot and $N = \{1, \dots, n\}$ denotes the customers. The distance between nodes $i$ and $j$ is $d_{ij}$, and $q_i$ is the demand of customer $i$, with $q_0 = 0$. A homogeneous fleet of vehicles is available, each with capacity $Q$; the number of vehicles used is not fixed in advance but determined by the solution. The objective is to serve all customers with a set of routes, each starting and ending at the depot, such that every route respects the vehicle capacity and the total travelled distance of all routes is minimized.

In the capacitated vehicle routing problem with time windows (CVRPTW), each customer $i$ additionally has a service time $p_i \geq 0$ and a time window $[a_i,b_i]$ during which service must start, and the depot imposes a scheduling horizon $[0,H]$ within which every vehicle must return. A vehicle may arrive at a customer before $a_i$ and wait, but service cannot start before $a_i$ nor after $b_i$.  Setting $p_i = 0$ and $[a_i,b_i] = [0,H]$ for all $i$, with a sufficiently large $H$, recovers the plain CVRP as a special case.

Introducing a binary decision variable $x_{ij}$, which assumes $1$ if a vehicle travels directly from node $i$ to node $j$ and $0$ otherwise, both problems can be captured by a single compact arc-based formulation. Degree and depot-balance constraints fix the route structure, a continuous load variable $u_i$ enforces the capacity restriction and eliminates subtours, and an optional continuous time variable $t_i$, present only when time windows apply, propagates schedules along each route:
\begin{align}
            & \min \sum_{\substack{(i,j) \in V^2 \\ i \neq j}} d_{ij} x_{ij},                    &  &                                       \label{eq:obj} \\
\text{s.t.} & \sum_{j \in V\setminus\{i\}} x_{ij} = \sum_{j \in V\setminus\{i\}} x_{ji} = 1,      &  & \forall i \in N,                      \label{eq:inout} \\
            & \sum_{j \in N} x_{0j} = \sum_{j \in N} x_{j0},                                     &  &                                       \label{eq:depot} \\
            & u_j \geq u_i + q_j - Q(1-x_{ij}),                                                  &  & \forall i,j \in N,\ i \neq j,          \label{eq:mtz} \\
            & q_i \leq u_i \leq Q,                                                               &  & \forall i \in N,                       \label{eq:load} \\
            & t_j \geq t_i + p_i + d_{ij} - M_{ij}(1-x_{ij}),                                    &  & \forall i,j \in N,\ i \neq j,          \label{eq:tw1} \\
            & t_j \geq d_{0j} - M_{0j}(1-x_{0j}),                                                &  & \forall j \in N,                       \label{eq:tw2} \\
            & t_i + p_i + d_{i0} \leq H + M_{i0}(1-x_{i0}),                                      &  & \forall i \in N,                       \label{eq:tw3} \\
            & a_i \leq t_i \leq b_i,                                                             &  & \forall i \in N,                       \label{eq:tw4} \\
            & x_{ij} \in \{0, 1\},
            & & \forall i,j \in
            V,\ i\neq j.                  
\label{eq:bin}
\end{align}
Here, \eqref{eq:obj} minimizes total distance; \eqref{eq:inout} ensures that each customer is entered and left exactly once; \eqref{eq:depot} balances the number of vehicles leaving and returning to the depot, leaving the fleet size free; \eqref{eq:mtz}--\eqref{eq:load} are Miller--Tucker--Zemlin-style load-propagation constraints, where $u_i$ tracks the cumulative load carried up to and including customer $i$, so that they enforce capacity feasibility along each route and, being strictly increasing along any used arc, also rule out subtours that do not include the depot; and \eqref{eq:bin} enforces integrality. Constraints \eqref{eq:tw1}--\eqref{eq:tw4}, included only for CVRPTW instances, are the time-window analogue of \eqref{eq:mtz}--\eqref{eq:load}: the variable $t_i$ denotes the service start time at customer $i$, propagated forward along each route by \eqref{eq:tw1}, with \eqref{eq:tw2} handling the first leg from the depot and \eqref{eq:tw3} the return leg, while \eqref{eq:tw4} restricts $t_i$ to the customer's time window. The constants $M_{ij}$ are big-M values, e.g., $M_{ij} = H + \max_k p_k + \max_{k,l} d_{kl}$, chosen large enough to deactivate the corresponding constraint whenever $x_{ij} = 0$.

Because \eqref{eq:mtz}--\eqref{eq:load} and \eqref{eq:tw1}--\eqref{eq:tw4} both replace an exponential family of cuts by a compact set of big-M constraints, this formulation is polynomial in size for either problem. Its linear relaxation is, however, considerably weaker than that of the classical rounded-capacity-cut formulations solved via branch-and-cut~\citep{toth2014vrp}; here, it serves primarily to give a precise, self-contained problem definition and to provide exact baselines at the small and medium instance sizes considered for the learning approach, rather than as a candidate for large-scale exact solving.

\section{Related work}\label{sec:related}

There is a large body of work on learning to solve combinatorial optimization problems, including constructive, search-based, and end-to-end approaches. A recent survey is given by \citet{Chung.etal/2025}. A related direction explores large language models (LLMs) as end-to-end solvers that map a natural-language problem description directly to a solution, bypassing intermediate code generation or solver calls. \citet{Jiang.etal/2025} fine-tune a $7$B-parameter LLM, first by supervised learning and then by reinforcement learning, to generate solutions for seven combinatorial problems. On graph instances with at most 100 nodes and scheduling instances with at most $20\times20$ operations, best-of-eight decoding attains average gaps of $1.03$--$8.20\%$, outperforming much larger general-purpose LLMs and several lightweight constructive heuristics in solution quality, but requiring longer inference times than the latter.

Among these specialized, reinforcement-learning-trained architectures, \emph{constructive} solvers build a route incrementally, one node at a time. For routing problems, this line of work goes back to sequence-to-sequence learning and pointer networks~\citep{Vinyals.etal/2015}, which \citet{Bello.etal/2016} first paired with a REINFORCE-trained policy gradient to construct tours directly from tour length, without labelled optimal solutions, and includes reinforcement-learning approaches for the CVRP such as \citet{Nazari.etal/2018}.

\citet{Kool.etal/2018} replace this recurrent pointer-network architecture with a fully attention-based encoder-decoder: an encoder embeds all nodes in parallel through several self-attention and feed-forward layers, and a decoder attends, from a single context vector updated as the tour is built, to these node embeddings to select the next node at each step, with the whole model trained end-to-end using REINFORCE and a greedy-rollout baseline.

The attention model (AM) has inspired several variants. \citet{Lei.etal/2021} add edge embeddings and residual connections in a residual edge-graph attention network, shifting a larger share of the computation into the encoder. \citet{Luo.etal/2023} move complexity into the decoder instead, with a ``light encoder, heavy decoder'' (LEHD) design whose heavier multi-layer decoder re-embeds only the first and last node of the partial tour at each step, letting a model trained on small instances generalize to ones an order of magnitude larger. \citet{Fang.etal/2024} restrict each decision to several fixed-size $k$-nearest-neighbour views and normalize their coordinates, so that a model trained at $n=100$ never processes the complete large instance at once. They evaluate this approach up to $n=10{,}000$ for TSP and $n=5{,}000$ for CVRP. \citet{Chen.etal/2025} retain this local decomposition and optimize the coordinate projection used at test time, reporting experiments with up to $100{,}000$ nodes. In the latter experiments, each neural call considers only $k=100$ nearby nodes; the reported scale refers to the surrounding instance, which is still traversed through $O(n)$ sequential decisions. Thus, scalability comes primarily from imposing a fixed local candidate set, while the learned model ranks candidates within that restricted view. These variants push complexity to opposite ends of the model: \citet{Lei.etal/2021} into the encoder, the LEHD family into the decoder. Since the QNN replacement studied here specifically targets the encoder, the encoder-light, decoder-heavy LEHD family is harder to adapt to it, and \citeauthor{Lei.etal/2021}'s heavier encoder would make the replacement more costly still---which is why this paper builds on the original, more balanced AM architecture instead.

A separate line of work keeps the constructive setup but targets constraints for which the usual feasibility mask is itself hard to compute. \citet{Bi.etal/2024} study routing variants with time windows and draft limits, where checking whether a candidate move keeps the rest of the route feasible is NP-hard in general; rather than an exact mask, they score violations with a Lagrangian penalty term and add a proactive infeasibility-prevention (PIP) step that looks one move ahead and forbids actions that would make an unvisited customer infeasible later in the same route, with a learned variant (PIP-D) that keeps this lookahead cheap. \citet{Bi.etal/2026} build on this with a lightweight refinement stage: a handful of learned local-search moves repair remaining violations after construction, and successful repairs are then fed back as imitation targets so the constructor itself gradually absorbs them. Because we assume an unlimited homogeneous fleet, a customer that cannot be added to the current route can still be served by a newly opened route. The specific PIP mechanism of \citet{Bi.etal/2024} is therefore unnecessary in our setting. Their broader point that some constraints are cheap to mask while others require dedicated architectural or training support is a useful complement to the parameter-efficiency question addressed in this paper.

A complementary family of methods decouples solution construction from a learned edge heatmap, using the heatmap to guide a generic search procedure such as Monte Carlo tree search. These methods differ mainly in how that heatmap is obtained: supervised graph-convolutional learning from sampled subgraphs~\citep{Fu.etal/2021}, REINFORCE learning of sparse edge preferences followed by active search~\citep{Qiu.etal/2022}, graph diffusion~\citep{Sun.Yang/2023}, or unsupervised optimization of a surrogate objective~\citep{Min.etal/2023}. All scale to instances with $n=10K$ nodes or more. However, \citet{Pan.etal/2024} show, in a systematic re-evaluation of this heatmap-plus-MCTS pipeline, that once the MCTS/$k$-opt search itself is properly tuned, the marginal contribution of the learned heatmap is small and inconsistent: at $n=10K$, the best result in their comparison has a relative deviation of only $2.13\%$ from the optimal tour length, and comes not from a learned model but from a fixed, parameter-free heatmap derived from nearest-neighbour rank. \citet{Xia.etal/2024} reach the same conclusion independently: their own parameter-free heatmap, a plain softmax over distances, already rivals or beats these learned ones, and the whole heatmap-plus-search paradigm is in turn beaten by the classical LKH-3 heuristic. This indicates that, in this family of methods, the strong instance-agnostic local search rather than the learned component accounts for most of the solution quality. Quantum-enhanced attention models have also been investigated for vehicle routing. \citet{Correll.etal/2023} integrate quantum orthogonal neural networks into the attention mechanism of a reinforcement-learning model for supply-chain logistics. In contrast, we investigate replacing encoder feed-forward modules and attention output projections with small variational QNNs, comparing them against width-matched classical bottlenecks for CVRP and CVRPTW.

\section{A hybrid quantum-classical neural network}\label{sec:hybrid}

Our approach takes inspiration from \citet{Kool.etal/2018}, who proposed a classical neural network for learning to route two-dimensional Euclidean instances. Their model consists of an encoder that takes the positions of the depot and customers, together with the customer demands, and produces a representation for each node. These representations are processed by several layers, each combining an attention mechanism, following the transformer architecture of \citet{Vaswani.etal/2017}, with a feed-forward layer. After the encoder, a decoder, which also employs an attention mechanism, iteratively extends partial routes until complete routes are obtained. For the CVRP, the approach produced good results on problem instances from $20$ up to $100$ customers, which covers the small-to-medium regime considered in this work.

\begin{figure}[!h]
\centering
\includegraphics[width=0.9\linewidth]{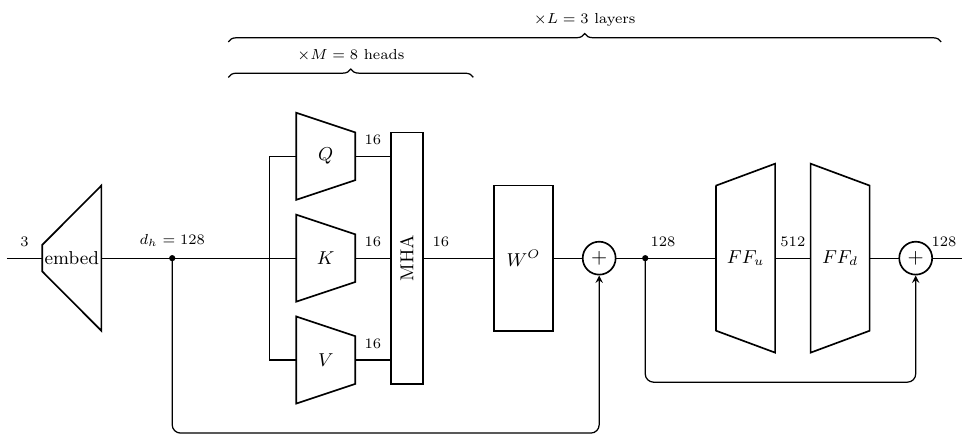}
\caption{Encoder architecture of the classical attention model. The figure shows the node embedding and one layer of a total of $L$ layers.}
\label{fig:encoder}
\end{figure}

\begin{figure}[tb]
\centering
\includegraphics[width=0.6\linewidth]{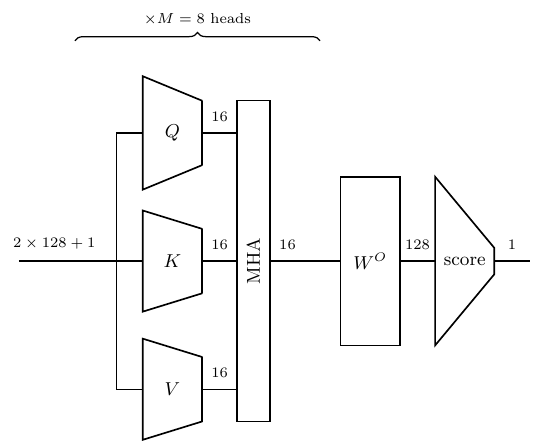}
\caption{Decoder architecture of the classical attention model.}
\label{fig:decoder}
\end{figure}

Figure~\ref{fig:encoder} shows the encoder and Figure~\ref{fig:decoder} shows the decoder. The encoder first embeds each node into a latent space of dimension $d_h$. Customers use their coordinates and demand, while the depot, whose demand is always zero, uses a separate coordinate projection:
\begin{equation}
h_i^{(0)} =
\begin{cases}
W^x [x_i\;y_i\;q_i]^\top + b^x, & i\neq 0,\\
W^0 [x_0\;y_0]^\top + b^0, & i=0.
\end{cases}
\label{eq:init-embed}
\end{equation}
followed by $L$ layers, each combining a multi-head attention (MHA) sublayer with $M=8$ heads and a feed-forward (FF) sublayer, both wrapped in a residual connection. Writing $d_k = d_h/M = 16$ for the per-head dimension, and $W_m^Q, W_m^K, W_m^V \in \mathbb{R}^{d_k\times d_h}$ and $W_m^O \in \mathbb{R}^{d_h\times d_k}$ for the per-head query, key, value, and output projections, $m=1,\dots,M$,
\begin{align}
h'_{im} &= \sum_{j \in V} \mathrm{softmax}_j\!\left(\frac{(W_m^Q h_i)^\top (W_m^K h_j)}{\sqrt{d_k}}\right) W_m^V h_j, \label{eq:attn}\\
\mathrm{MHA}_i(h) &= \sum_{m=1}^M W_m^O h'_{im}. \label{eq:mha}
\end{align}
Equivalently, the $M$ head outputs may be concatenated into $[h'_{i1};\ldots;h'_{iM}]\in\mathbb{R}^{d_h}$ and multiplied by the block output matrix $W^O=[W_1^O\;\cdots\;W_M^O]\in\mathbb{R}^{d_h\times d_h}$. This is the encoder attention output projection considered for replacement below. For each layer $\ell = 1,\dots,L$,
\begin{align}
\hat h_i^{(\ell)} &= \operatorname{BN}_1^{(\ell)}\!\left(h_i^{(\ell-1)} + \mathrm{MHA}_i\big(h^{(\ell-1)}\big)\right), \label{eq:res-mha}\\
h_i^{(\ell)} &= \operatorname{BN}_2^{(\ell)}\!\left(\hat h_i^{(\ell)} + FF\big(\hat h_i^{(\ell)}\big)\right), \label{eq:res-ff}
\end{align}
where $FF(x) = W_2\,\mathrm{ReLU}(W_1 x + b_1) + b_2$ is a two-layer feed-forward network with hidden dimension $d_{ff}=512$. Following \citet{Kool.etal/2018}, each residual addition is followed by an independent batch-normalization layer with learned scale and shift.

The encoder produces a final node embedding $h_i = h_i^{(L)}$ for $i\in V$ as well as a graph embedding $\overline h = \sum_{i\in V} h_i/|V|$.

The decoder builds a route by repeatedly attending from a single context query to the node embeddings. At step $t$, with $\pi_t$ the last visited node and $\rho_t$ the remaining vehicle capacity, the context query combines the graph embedding with the current state through two separate linear projections that are summed rather than concatenated,
\begin{equation}
q_c = W^Q_{\mathrm{graph}}\,\overline h
    + W^Q_{\mathrm{step}}
    \begin{bmatrix} h_{\pi_t} \\ \rho_t \end{bmatrix},
\label{eq:context}
\end{equation}
This is the CVRP-specific context of \citet{Kool.etal/2018}: because a route may be reopened from the depot at any time, only the last visited node and the remaining capacity are needed. The query is refined by one further application of the same multi-head attention operator as the encoder, attending over all node embeddings,
\begin{equation}
\tilde q_c = \mathrm{MHA}_c(q_c, h),
\label{eq:glimpse}
\end{equation}
and a final single-head compatibility layer produces masked, tanh-clipped logits over the candidate nodes,
\begin{equation}
u_j =
\begin{cases}
C \cdot \tanh\!\left(\dfrac{\tilde q_c^\top k_j}{\sqrt{d_k}}\right), & j \text{ feasible},\\[6pt]
-\infty, & \text{otherwise},
\end{cases}
\qquad
p_j = \frac{e^{u_j}}{\sum_{j'} e^{u_{j'}}},
\label{eq:logits}
\end{equation}
where $k_j = W^K h_j$ and $C=10$ clips the logits, following \citet{Kool.etal/2018}. Already visited customers, as well as customers whose demand would exceed the remaining capacity, are masked before the softmax. A route is constructed iteratively from $p$, either greedily or by sampling; when no feasible successor remains, a new route is opened from the depot. This is repeated until all customers are served.

For the CVRPTW variant, time windows are enforced as hard constraints via masking, not as a penalty in the objective. Writing $t$ for the vehicle's current time and $\pi_t$ for its current location, a candidate customer $j$ is masked as infeasible, in addition to already visited or over-capacity customers, whenever
\begin{equation}
\max(t+d_{\pi_t j},\,a_j) > b_j
\qquad\text{or}\qquad
\max(t+d_{\pi_t j},\,a_j) + p_j + d_{j0} > H,
\label{eq:tw-mask}
\end{equation}
i.e.\ whenever service could not start before the due date $b_j$, or serving $j$ would prevent the vehicle from returning to the depot before the horizon $H$ closes. A vehicle arriving before $a_j$ simply waits, so only late arrival, never early arrival, is restricted. Because the fleet is unlimited, masking a customer out of the current route costs nothing beyond ending that route: every instance is generated so that each customer is individually feasible for a direct depot--customer--depot trip (Section~\ref{sec:methodology}), so a fresh route from the depot can always serve whatever remains once the current route closes. A feasible complete solution is therefore always reached without ever violating a window, which is why hard masking, rather than the soft time-window penalties more common elsewhere in the literature (Section~\ref{sec:related}), is a natural fit for this unlimited-fleet setting.

\citet{Kool.etal/2018} propose a hidden size of $d_h=128$ and a hidden layer of size $512$ in the feed-forward network. Therefore, the sizes and evaluation counts per forward step of the components are as shown in Table~\ref{tab:components}. Note that the decoder does not use the usual all-to-all attention mechanism, but rather a one-to-all attention of the context node to all other nodes. This is for performance reasons, since the decoder step must be repeated $n+k$ times until a complete routing solution with $k$ vehicles is produced. Nevertheless, this remains the hottest path, requiring $O(|V|^2)$ attention evaluations over the construction of a complete solution.

\begin{table}[!h]
\centering
 \caption{Component sizes and number of evaluations during the construction of a complete solution in the classical attention model.}
\label{tab:components}
\begin{tabular}{lcc}
\toprule
Component           & \#parameters & \#evaluations    \\
\midrule
\multicolumn{3}{c}{Encoder} \\
\midrule
Initial embeddings              & 896            & $|V|$            \\
Attention queries, keys, values & $49{,}152\,L$  & $L|V|$           \\
Attention output $W^O$          & $16{,}384\,L$  & $L|V|$           \\
Feed-forward                    & $131{,}712\,L$ & $L|V|$           \\
Batch normalization             & $512\,L$     & $L|V|$           \\
\midrule
\multicolumn{3}{c}{Decoder} \\
\midrule
Attention layer   & $98,432$     & $|V|^2$          \\
\midrule
Totals            & $692{,}608$    & $O(|V|^2+L|V|)$ \\
\bottomrule
\end{tabular}
\end{table}

For $L=3$, the feed-forward layers contain $395{,}136$ parameters and therefore account for more than half of the model. The encoder attention output projections contribute another $49{,}152$ parameters. In particular, each feed-forward sublayer maps $128\to512\to128$ and contains $131{,}712$ parameters, whereas each $W^O$ maps the concatenated $128$-dimensional head representation back to $128$ dimensions and contains $16{,}384$ parameters. Thus $W^O$ is smaller than the feed-forward sublayer, rather than a $512\times512$ matrix, but it is still a repeated dense encoder component outside the iterative decoder path.

\begin{figure}[tb]
\centering
\includegraphics[width=\linewidth]{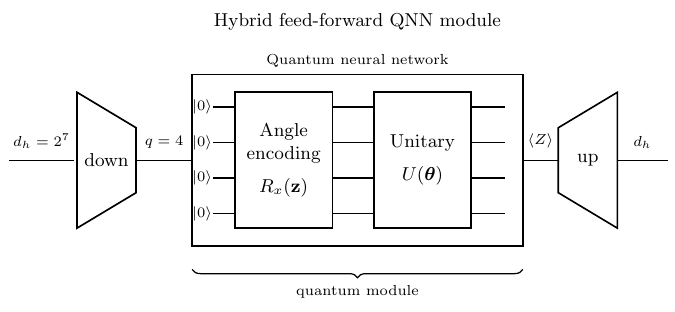}
\caption{Hybrid replacement of the encoder feed-forward block.}
\label{fig:qnn}
\end{figure}

We therefore study two replacement levels. \emph{Single compression} replaces every encoder feed-forward sublayer by a narrow classical or quantum $128\to q\to128$ bottleneck module. \emph{Double compression} additionally replaces every encoder attention output projection $W^O$ by a module with the same bottleneck. For each level we compare a classical bottleneck with a QNN bottleneck at matched width $q=4$. As illustrated in Figure~\ref{fig:qnn}, the QNN module first projects the $d_h$-dimensional input to $q$ values, which are encoded through $R_X$ rotations. A trainable unitary $U(\boldsymbol{\theta})$ implements a brickwall ansatz consisting of $R_X$, $R_Y$, and $R_Z$ rotations and CNOT entangling gates. The resulting Pauli-$Z$ expectation values are projected back to dimension $d_h$. Selecting $q\ll d_h$ imposes the desired strong bottleneck while allowing the same module to replace either dense component.

\section{Methodology}\label{sec:methodology}

\subsection{Problem Instances}
Training is performed on randomly generated Euclidean instances with varying numbers of nodes $n \in \{10, 20, 50, 100\}$, where node coordinates of the customers and the depot are sampled uniformly from the unit square $[0,1]^2$. Customer demands are drawn as random integers from $\{1,2,\ldots,9\}$, and vehicle capacities are set to $20$, $30$, $40$, and $50$ for instances with $n = 10$, $20$, $50$, and $100$ customers, respectively.

For the CVRPTW variant, the same underlying instances are extended with a per-customer service time and time window, following the classical distinction of \citet{Solomon/1987} into families R1 (tight time windows) and R2 (wide time windows). Every customer has a fixed service time of $0.1$, and the scheduling horizon is set to $H=3$ for R1 instances and $H=10$ for R2 instances; both values are rescaled from Solomon's original benchmarks to the unit square used here, and chosen large enough that every customer remains individually feasible for a direct depot--customer--depot round trip even when the depot falls near a corner. For each customer $i$, a window centre is drawn uniformly at random from the interval of times for which such a round trip remains feasible, and a window half-width is drawn from a folded normal distribution, $\lvert\mathcal{N}(0.1,0.05^2)\rvert$ for R1 and $\lvert\mathcal{N}(0.5,0.25^2)\rvert$ for R2, following Solomon's original construction; the half-width is lower-bounded at $10^{-3}$ to avoid degenerate windows and then capped by the largest individually feasible half-width, so that the resulting window $[a_i,b_i]$ stays within $[0,H]$.

\subsection{Models and Training Setup}

All training and evaluation experiments were performed on a workstation equipped with an Intel Core i7-13700KF processor, 32~GiB of RAM, and an NVIDIA GeForce RTX~3060 GPU with 12~GiB of memory. Neural training and evaluation use CUDA, while the classical routing solvers use Release-mode CPU implementations.

Both the classical and hybrid neural networks are trained using a batch size of $2^7$ samples and $2^8$ batches per epoch, resulting in an epoch size of $2^{15}$ samples, and a total of $2^7$ training epochs. This setup deviates from the training regime of \citet{Kool.etal/2018} in two respects. First, the number of samples per epoch is reduced by approximately a factor of $40$ compared to the $1{,}280\mathrm{K}$ samples used there. Second, preliminary experiments indicate that improved performance is often achieved in later epochs, motivating an increase in the total number of training epochs.

Following \citet{Kool.etal/2018}, the policy is trained end-to-end using the REINFORCE algorithm \citep{Williams/1992}, combined with a greedy rollout baseline, which demonstrated superior performance in prior work. Optimization is carried out using the Adam optimizer with a learning rate of $10^{-4}$. The QNN components are simulated and trained via backpropagation. A compact QNN configuration is adopted, consisting of $q = 4$ qubits and $L_q = 2$ layers. Each QNN bottleneck contains $1{,}024$ classical projection weights and $24$ trainable circuit parameters. With $L=3$ encoder layers, single compression therefore contains $300{,}544$ classical and $72$ quantum parameters, for $300{,}616$ trainable parameters in total. This is a reduction of $391{,}992$ ($56.6\%$) from the $692{,}608$-parameter uncompressed model. Double compression uses six QNN bottlenecks and contains $254{,}464$ classical and $144$ quantum parameters, or $254{,}608$ in total, reducing the original model by $438{,}000$ parameters ($63.2\%$). The corresponding matched classical single- and double-compression models contain $300{,}940$ and $254{,}860$ parameters, respectively.

 Each model in the broad comparison is trained once, using training seed 1234, following a curriculum over $n=10\to20\to50\to100$. At each size, the checkpoint with the lowest objective on a fixed set of $1{,}000$ validation instances generated with seed 4321 is selected. Its model and Adam state initialize the next size, while the rollout baseline is rebuilt for the new instance size. Model selection never uses the held-out test instances.  Due to the computational cost of training the simulated QNNs, we use a single training seed for each configuration. Consequently, the comparisons characterize performance under a fixed training protocol rather than variability across independent training runs.

The selected models are evaluated using both greedy decoding and stochastic sampling on a fixed held-out set of $N = 1{,}000$ instances generated with seed 9876. For the sampling-based evaluation, the best solution among $1{,}280$ samples is reported.

\subsection{Baseline Algorithms}

Performance is compared against GOT, the routing solver from Google OR-Tools v9.15 \citep{ortools_routing}, and LKH3, the Lin--Kernighan--Helsgaun heuristic \citep{Helsgaun/2000,Helsgaun/2017}. For plain CVRP, we reproduce the LKH3 protocol released by \citet{Kool.etal/2018}. Because that implementation leaves the number of vehicles unspecified, LKH3 fixes it to the aggregate-demand lower bound $K=\lceil\sum_i q_i/Q\rceil$. This baseline therefore solves a fixed-minimum-fleet restriction of our unlimited-fleet CVRP. We use one run with $1{,}000$ trials per instance.

For CVRPTW we additionally use Solomon-I1, the classical sequential-insertion heuristic of \citet{Solomon/1987}. Each route starts from one unrouted seed customer. For a feasible insertion of customer $u$ between consecutive nodes $i$ and $j$, I1 combines the added distance $c_{11}=d_{iu}+d_{uj}-d_{ij}$ with the resulting delay $c_{12}$ at the successor into $c_1=\alpha c_{11}+(1-\alpha)c_{12}$. For each customer, the insertion position minimizing $c_1$ is retained, and the customer maximizing $c_2=\lambda d_{0u}-c_1$ is inserted. Thus $\alpha$ trades distance against schedule disruption, while $\lambda$ controls the preference for customers far from the depot. When no further feasible insertion exists, the route is closed and a new seed is selected. We run all combinations of $\alpha\in\{1,0.75,0.5\}$, $\lambda\in\{1,2\}$, and either farthest-depot or earliest-due-date seeding, retaining the best of the resulting $12$ solutions. Since selecting an unrestricted fleet size directly with LKH3 is not scalable, our CVRPTW hybrid first obtains a feasible GOT solution with $K_{\mathrm{GOT}}$ routes, then runs LKH3 for $1{,}000$ trials at each of $K_{\mathrm{GOT}}$, $K_{\mathrm{GOT}}+1$, and $K_{\mathrm{GOT}}+2$. The best valid solution among these three runs and the original GOT solution is reported as GOT--LKH3.

\section{Experimental results}\label{sec:results}
This section reports computational results comparing the proposed learning-based approach with classical algorithms. All reported solutions are explicitly checked for feasibility. For the neural models, decoder masking enforces feasibility by construction, and no invalid routes were observed in our experiments.

Table~\ref{tab:results} reports the matched single- and double-compression models under sample-$1{,}280$ decoding together with the classical baselines. Every panel uses the same $1{,}000$ seed-9876 held-out instances for all methods. ``Obj.'' is mean route length, $\bar r$ is mean instance-wise relative deviation, and $t$ is total evaluation time in minutes.

For each problem family and size, let $v_{im}$ denote the objective obtained by method $m$ on instance $i$, let $I_m$ be the instances on which that method returns a valid solution, and let $\mu_i=\min_m v_{im}$ over the valid results in the table. We report $\bar r_m=100|I_m|^{-1}\sum_{i\in I_m}(v_{im}/\mu_i-1)$. Thus each relative-deviation pool contains exactly the methods displayed in the corresponding panel.

\begin{table*}[!h]
\centering
\caption{Held-out CVRP and CVRPTW results. Objective values are route lengths in the unit square; $\bar r$ is the mean relative deviation from the best displayed method per instance; and $t$ is total evaluation time in minutes. Neural rows use the best of $1{,}280$ samples. GOT--LKH3 reports the best of GOT and three 1,000-trial LKH3 runs at $K_{\mathrm{GOT}},K_{\mathrm{GOT}}+1,K_{\mathrm{GOT}}+2$. $^\dagger$At CVRP $n=10$, fixed-minimum-fleet LKH3 returned 993 valid solutions; all other entries contain 1,000.}
\label{tab:results}
\begingroup
\setlength{\tabcolsep}{2.2pt}
\scriptsize
\resizebox{\textwidth}{!}{%
\begin{tabular}{lrrrrrrrrrrrr}
\toprule
& \multicolumn{3}{c}{$n=10$} & \multicolumn{3}{c}{$n=20$} & \multicolumn{3}{c}{$n=50$} & \multicolumn{3}{c}{$n=100$} \\
\cmidrule(lr){2-4}\cmidrule(lr){5-7}\cmidrule(lr){8-10}\cmidrule(lr){11-13}
Method & Obj. & $\bar r$ (\%) & $t$ (min) & Obj. & $\bar r$ (\%) & $t$ (min) & Obj. & $\bar r$ (\%) & $t$ (min) & Obj. & $\bar r$ (\%) & $t$ (min) \\
\midrule
\multicolumn{13}{l}{\textit{CVRP}} \\
Classical (single) & 4.639 & 1.95 & 0.2 & 6.370 & 2.44 & 0.4 & 10.776 & 3.36 & 1.4 & 16.556 & 4.82 & 4.2 \\
QNN (single) & 4.623 & 1.65 & 0.3 & 6.370 & 2.45 & 0.5 & 10.810 & 3.67 & 1.5 & 16.639 & 5.33 & 4.3 \\
Classical (double) & 4.625 & 1.67 & 0.2 & 6.362 & 2.32 & 0.4 & 10.810 & 3.69 & 1.4 & 16.630 & 5.31 & 4.2 \\
QNN (double) & 4.615 & 1.47 & 0.4 & 6.370 & 2.44 & 0.6 & 10.841 & 3.99 & 1.6 & 16.747 & 6.04 & 4.4 \\
GOT & 4.650 & 2.31 & $<0.1$ & 6.523 & 4.95 & 0.1 & 11.270 & 8.13 & 0.3 & 17.242 & 9.23 & 0.9 \\
LKH3 (1K)$^\dagger$ & 4.561 & 0.38 & 1.9 & 6.221 & 0.06 & 7.1 & 10.428 & 0.01 & 22.5 & 15.797 & 0.00 & 40.2 \\
\midrule
\multicolumn{13}{l}{\textit{CVRPTW--R1}} \\
Classical (single) & 5.903 & 1.41 & 0.3 & 9.413 & 2.28 & 0.5 & 17.927 & 4.64 & 1.7 & 29.316 & 6.97 & 5.0 \\
QNN (single) & 5.903 & 1.42 & 0.4 & 9.432 & 2.48 & 0.6 & 17.898 & 4.46 & 1.8 & 29.374 & 7.17 & 5.2 \\
Classical (double) & 5.897 & 1.32 & 0.3 & 9.412 & 2.28 & 0.5 & 17.927 & 4.66 & 1.7 & 29.532 & 7.76 & 5.0 \\
QNN (double) & 5.896 & 1.32 & 0.5 & 9.413 & 2.31 & 0.7 & 18.036 & 5.31 & 1.9 & 29.702 & 8.40 & 5.2 \\
GOT & 5.886 & 1.19 & $<0.1$ & 9.325 & 1.29 & 0.1 & 17.609 & 2.71 & 0.3 & 28.367 & 3.40 & 1.4 \\
GOT--LKH3 (1K) & 5.828 & 0.11 & 7.6 & 9.214 & 0.09 & 14.1 & 17.153 & 0.06 & 28.4 & 27.440 & 0.03 & 50.1 \\
Solomon-I1 & 5.967 & 2.70 & $<0.1$ & 9.630 & 4.84 & $<0.1$ & 18.506 & 8.27 & $<0.1$ & 30.108 & 10.13 & 0.1 \\
\midrule
\multicolumn{13}{l}{\textit{CVRPTW--R2}} \\
Classical (single) & 5.017 & 2.75 & 0.2 & 7.499 & 4.93 & 0.5 & 13.722 & 7.64 & 1.6 & 21.791 & 10.36 & 4.7 \\
QNN (single) & 5.030 & 3.05 & 0.4 & 7.476 & 4.61 & 0.6 & 13.669 & 7.22 & 1.7 & 21.804 & 10.43 & 4.8 \\
Classical (double) & 5.001 & 2.47 & 0.2 & 7.452 & 4.28 & 0.5 & 13.639 & 6.99 & 1.6 & 21.902 & 10.94 & 4.7 \\
QNN (double) & 5.011 & 2.65 & 0.5 & 7.441 & 4.12 & 0.7 & 13.752 & 7.89 & 1.8 & 22.115 & 12.02 & 4.9 \\
GOT & 5.058 & 3.66 & $<0.1$ & 7.601 & 6.39 & 0.1 & 14.179 & 11.21 & 0.3 & 22.442 & 13.66 & 1.0 \\
GOT--LKH3 (1K) & 4.882 & 0.03 & 9.7 & 7.146 & 0.01 & 13.7 & 12.755 & 0.05 & 32.4 & 19.757 & 0.04 & 63.3 \\
Solomon-I1 & 5.241 & 7.66 & $<0.1$ & 8.205 & 15.03 & $<0.1$ & 15.427 & 21.20 & $<0.1$ & 24.277 & 23.15 & 0.2 \\
\bottomrule
\end{tabular}%
}
\endgroup
\end{table*}

We begin with an examination of overall expected objective values. \citet{Haimovich.RinnooyKan/1985} give an asymptotic estimate for the expected optimal CVRP tour length, combining the cost of returning to the depot with the cost of local routing among customers:
\begin{equation}
L_{\mathrm{CVRP}} \approx \frac{2n\bar q}{Q}\,\bar d_0 + \beta_2\sqrt{n},
\label{eq:cvrp-estimate}
\end{equation}
where $\bar q$ is the expected customer demand, $\bar d_0$ is the expected distance between the depot and a customer, and $\beta_2 \approx 0.71$ is the Beardwood--Halton--Hammersley constant~\citep{Beardwood.etal/1959}. For our instances, both the depot and the customers are drawn uniformly from the unit square, so the expected distance between them is $\bar d_0 \approx 0.5214$. With $\bar q = 5$ and capacities $20$, $30$, $40$, and $50$ for $n=10$, $20$, $50$, and $100$, respectively, Equation~\eqref{eq:cvrp-estimate} yields estimates of $4.85$, $6.65$, $11.54$, and $17.53$.  These track the objective values in Table~\ref{tab:results} closely across all four instance sizes, suggesting that the observed CVRP objective values lie within the expected range.

At $n=10$ and $20$, the matched classical and quantum bottlenecks differ by at most $0.35\%$, without a consistent ordering. Single compression remains within approximately 0.5\% at every size, with differences changing sign across problem families and sizes. A clearer size-dependent pattern appears under double compression: at $n=50$ and $100$, the QNN objective is higher in all six comparisons, by $0.29\%$--$0.97\%$.  Thus, replacing only the encoder feed-forward blocks preserves the quality of the matched classical bottleneck more consistently, whereas also replacing the attention output projections produces a small but consistent degradation in the reported experiments at larger sizes. The additional replacement increases the parameter reduction from $56.6\%$ to $63.2\%$, but does not improve solution quality.

To assess the robustness of the central single-compression CVRP comparison, we repeated the complete $n=10\to20\to50\to100$ curriculum for five training seeds. Under sample-$1{,}280$ decoding, the paired QNN-minus-classical differences in mean route length range from $-0.05\%$ to $+0.15\%$ across sizes; every seed-level difference is within $0.54\%$. Thus, the practical parity between the feed-forward QNN and its matched classical bottleneck is robust across these five training seeds. This replication concerns CVRP single compression only; the CVRPTW and double-compression rows in Table~\ref{tab:results} remain single-seed comparisons.

The specialized classical methods provide a stronger solution-quality reference. Fixed-minimum-fleet LKH3 has the smallest mean CVRP objective at every size. On CVRPTW, the GOT--LKH3 protocol dominates the mean objective and is within $0.12\%$ mean relative deviation of the per-instance pool minimum in every R1 and R2 column. GOT alone is substantially faster but weaker, while Solomon-I1 degrades most rapidly with size. These comparisons do not support solver-dominance or quantum-advantage claims; they delimit the quality retained by the compressed learned policies.

In runtime terms, GOT and Solomon-I1 are inexpensive, neural sample-$1{,}280$ decoding takes seconds to a few minutes per test set, and the multi-start GOT--LKH3 protocol is deliberately more computationally intensive in exchange for stronger solutions.

Taken together, these results give a qualified positive answer to the central design question: a four-qubit QNN can replace the encoder feed-forward layers while preserving the performance of the corresponding narrow classical model to within approximately $0.5\%$, reducing the original model by $56.6\%$. Extending the replacement to the attention output projections increases the reduction to $63.2\%$, but is less robust at larger sizes. Crucially, the matched classical bottlenecks achieve essentially the same quality. Most of the parameter reduction therefore comes from the narrow bottleneck itself rather than from quantum processing; the result demonstrates a viable quantum replacement, not a quantum-specific compression advantage.

\section{Conclusion}\label{sec:conclusion}

We asked whether small QNN bottlenecks can replace parameter-heavy modules in an attention-based routing policy. Our answer is a qualified yes. Replacing the encoder feed-forward layers with four-qubit QNNs reduces the model by $56.6\%$, while their objectives remain within approximately $0.5\%$ of those of the corresponding narrow classical models across all tested problem families and sizes. Replacing the attention output projections as well increases the reduction to $63.2\%$, but produces a consistent degradation at larger sizes in the reported experiments. Single compression is therefore the better-supported design.

The central qualification is that matched classical bottlenecks perform equally well while containing nearly the same total number of parameters. The observed compression consequently comes primarily from imposing a narrow intermediate representation, not from quantum processing. Our results establish that a compact QNN can serve as a viable module within a routing policy, but they do not demonstrate a quantum-specific performance or compression advantage. Specialized classical routing methods also remain stronger solution-quality references.

This outcome is consistent with broader QML evidence that practical advantages remain scarce and often weaken under carefully tuned, capacity-matched classical controls~\citep{Schuld.Petruccione/2021,HibatAllah.etal/2023,Jerbi.etal/2021,Peters.etal/2021,Franz.etal/2024,Bowles.etal/2024}. Rather than being exceptional, the present result extends that evidence to a matched comparison of compressed routing policies.

Because the QNN components are simulated, these results concern model quality and parameterization rather than execution speed or resource advantages on quantum hardware. The next question is whether quantum-specific properties---such as inductive bias, sample efficiency, or hardware execution---can provide a benefit unavailable to an equally compact classical component.

\backmatter

\bmhead{Acknowledgments}

This work was executed under the TIC26 -- Brazil Quantum Camp project, funded within the scope of the Prioritized Informatics Programs and Projects (PPI), Process No. 01245.008254/2025-22, under the responsibility of the Ministry of Science, Technology and Innovation (MCTI), with operational coordination by the Association for the Promotion of Brazilian Software Excellence (SOFTEX), and executed by CESAR and the Instituto de Pesquisas Eldorado. The authors also want to thank VCI Vanguard Confec\c{c}\~oes Importadas S.A.~(Aramis) for their collaboration on solving practical vehicle routing problems with hybrid approaches.


\section*{Declarations}

\bmhead{Data Availability}

The CVRP and CVRPTW instances used in this study were
synthetically generated following the procedures described
in Section~\ref{sec:methodology}. The generated instances
and experimental results are not publicly available due to
confidentiality restrictions of the Instituto de Pesquisas
Eldorado, but may be obtained from the corresponding author
on reasonable request, subject to institutional approval.

\bmhead{Code Availability}

The source code and scripts developed for this study are
proprietary to the Instituto de Pesquisas Eldorado and are
not publicly available. Further details may be obtained
from the corresponding author on reasonable request,
subject to institutional approval.

\bmhead{Competing Interests}

The authors declare no competing interests.

\bibliography{sn-bibliography}

\end{document}